\documentclass[conference]{IEEEtran}
\IEEEoverridecommandlockouts

\usepackage[utf8]{inputenc}
\usepackage[T1]{fontenc}

\usepackage{cite}
\usepackage{amsmath,amssymb,amsfonts}
\usepackage{algorithmic}
\usepackage{subfig}
\usepackage{graphicx}
\usepackage{textcomp}
\usepackage{xcolor}
\usepackage{booktabs}
\usepackage{multirow}
\usepackage{hyperref}

\usepackage[finalnew]{trackchanges}
\usepackage{subcaption}
\def\BibTeX{{\rm B\kern-.05em{\sc i\kern-.025em b}\kern-.08em
    T\kern-.1667em\lower.7ex\hbox{E}\kern-.125emX}}
\begin{document}

\title{Detection-segmentation convolutional neural network for autonomous vehicle perception}


\author{
\IEEEauthorblockN{Maciej Baczmanski, Robert Synoczek, Mateusz Wasala, Tomasz Kryjak}
\IEEEauthorblockA{Embedded Vision Systems Group, Department of Automatic Control and Robotics \\
AGH University of Krakow, Poland \\
\{mbaczmanski, synoczek\}@student.agh.edu.pl, 
\{mateusz.wasala, tomasz.kryjak\}@agh.edu.pl} \\
}

\maketitle

\begin{abstract}

Object detection and segmentation are two core modules of an autonomous vehicle perception system.
They should have high efficiency and low latency while reducing computational complexity.
Currently, the most commonly used algorithms are based on deep neural networks, which guarantee high efficiency but require high-performance computing platforms.
In the case of autonomous vehicles, i.e. cars, but also drones, it is necessary to use embedded platforms with limited computing power, which makes it difficult to meet the requirements described above.
A reduction in the complexity of the network can be achieved by using an appropriate: architecture, representation (reduced numerical precision, quantisation, pruning), and computing platform.
In this paper, we focus on the first factor -- the use of so-called detection-segmentation networks as a component of a perception system.
We considered the task of segmenting the drivable area and road markings in combination with the detection of selected objects (pedestrians, traffic lights, and obstacles).
We compared the performance of three different architectures described in the literature: MultiTask V3, HybridNets, and YOLOP.
We conducted the experiments on a custom dataset consisting of approximately 500 images of the drivable area and lane markings, and 250 images of detected objects.
Of the three methods analysed, MultiTask V3 proved to be the best, achieving 99\% $mAP_{50}$ for detection, 97\% MIoU for drivable area segmentation, and 91\% MIoU for lane segmentation, as well as 124 fps on the RTX 3060 graphics card.
This architecture is a good solution for embedded perception systems for autonomous vehicles.
The code is available at: \url{https://github.com/vision-agh/MMAR_2023}.

\end{abstract}

\begin{IEEEkeywords}
detection-segmentation convolutional neural network, autonomous vehicle, embedded vision, YOLOP, HybridNets, MultiTask V3
\end{IEEEkeywords}




\begin{figure}[!]
\centerline{\includegraphics[width=0.4\textwidth]{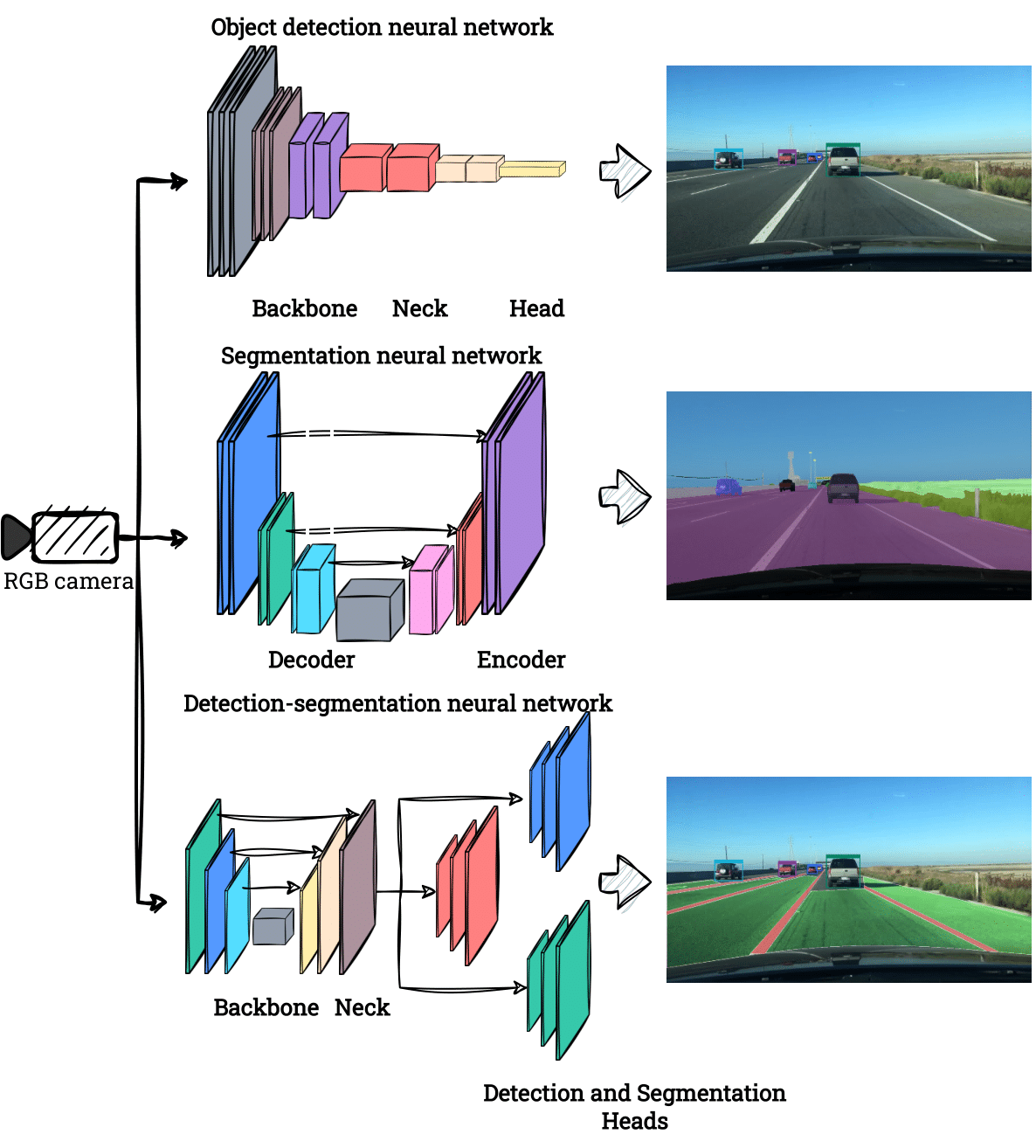}}
\caption{Illustration of the discussed network architectures.}
\label{fig:photoabstract}
\end{figure}

\section{Introduction}
\label{sec:intro}



\change{Perception systems are used in a wide range of mobile robots, including self-driving cars, advanced driver assistance systems (ADAS), and unmanned aerial vehicles (UAV -- often referred to as drones).
Based on data from various sensors: cameras (video, NIR (Near Infrared), thermal, event), radar, ultrasonic sensors, LiDAR (Light Detection and Ranging), IMU (Inertial Measurement Unit), and GNSS (Global Navigation Satellite Systems), they provide the control system with information about the vehicle's position in 3D space and all potential objects relevant to the task of planning and executing the vehicle's trajectory (for cars, e.g. other cars, pedestrians, cyclists, traffic lights, etc.).
}{Perception systems in mobile robots, including self-driving cars and unmanned aerial vehicles (UAV), use sensors like cameras, LiDAR (Light Detection and Ranging), radar, IMU (Inertial Measurement Unit), GNSS (Global Navigation Satellite Systems) and more to provide crucial information about the vehicle's position in 3D space and detect relevant objects (e.g. cars, pedestrians, cyclists, traffic lights, etc.).}
\change{When processing image and LiDAR sensor data, two tasks can be distinguished -- detection and segmentation (semantic, instance).
The first is based on detecting a selected class of objects (e.g. cars, pedestrians, obstacles) and labeling them in the form of a bounding box and sometimes an object mask.
On the other hand, semantic segmentation involves assigning an appropriate label to each pixel, depending on what it represents in the image (e.g. objects, drivable area, road markings, grass, vegetation, etc.).
Instance segmentation assigns different labels to objects belonging to the same class (e.g. different cars). 
This allows all objects to be correctly identified and tracked.
}{Image and LiDAR data processing involve two main tasks: detection, which identifies objects and labels them with bounding boxes or masks, and segmentation, which assigns labels to each pixel based on its representation in the image. Instance segmentation assigns different labels to objects belonging to the same class (e.g. different cars). This allows all objects to be correctly identified and tracked.}
Typically, both tasks are performed by different types of deep convolutional neural networks. 
For detection, networks from the YOLO family (\textit{You Only Look Once} \cite{yolov7}) are the most commonly used solution.
For segmentation, networks based on the CNN architecture are used, such as U-Net \cite{unet} and the fully convolutional network for semantic segmentation, and the mask R-CNN for instance segmentation. 
It is also worth mentioning the increasing interest in transformers neural networks in this context \cite{transformers_seg}. 
However, the use of two independent models has a negative impact on the computational complexity and energy efficiency of the system.
For this reason, network architectures that perform both of the above tasks simultaneously are being researched.
There are two approaches that can be used to solve this challenge: using instance segmentation networks or detection-segmentation networks. 
Instance segmentation networks are a special class of segmentation networks and require the preparation of a training dataset that is common to all detected objects. 
In addition, their operation is rather complex, and only part of the results are used for self-driving vehicles (distinguishing instances of classes such as road, lane, etc. is unnecessary for further analysis and often difficult to define precisely). 
Detection-segmentation networks consist of a common part (called the backbone) and several detection and segmentation heads.
This architecture allows the preparation of a separate training dataset for detection and often several subsets for segmentation (e.g. a separate one for lane and road marking segmentation).
This allows the datasets to be scaled according to how important the accuracy of the module is. 
In addition, the datasets used can contain independent sets of images, which greatly simplifies data collection and labeling.
The three architectures discussed so far: detection, segmentation, and detection-segmentation are shown in Figure \ref{fig:photoabstract}.
%
%
In addition, limiting the number of classes will reduce the time needed for post-processing, which involves filtering the resulting detections, e.g. using the NMS (Non-Maxima Suppression) algorithm. 
\change{In addition, s}{S}egmenting the image into only selected categories can also reduce inference time and increase accuracy.
All these arguments make detection-segmentation networks a good solution for embedded perception systems for autonomous vehicles.

In this paper, we compared the performance of three detection-segmentation networks: MultiTask V3  \cite{multitask}, HybridNets  \cite{HybridNet}, and YOLOP \cite{YOLOP}.
We conducted the experiments on a~custom dataset, recorded on a mock-up of a city.
The road surface and road markings were segmented, and objects such as pedestrians, traffic lights, and obstacles were detected.
To the best of our knowledge, this is the first comparison of these methods presented in the scientific literature.




The rest of the paper is structured as follows. 
Section \ref{sec:related_works} discusses the most important works on the use of neural networks for simultaneous object detection and segmentation. 
The architectures of the tested networks are then presented in Section \ref{sec:det_seg_nn}.
The methods for training the neural networks are described in Section \ref{sec:experiments}.
The results obtained are summarised in Section \ref{sec:results}. 
The paper ends with conclusions and a discussion of possible future research.

\section{Related works}
\label{sec:related_works}


Many different methods have been described in the scientific literature for the detection of drivable area and road markings, as well as for the detection of objects, e.g. pedestrians, cars, traffic signs, traffic lights, etc.
One of the solutions available is the use of deep neural networks. 
These can be divided into detection, segmentation, and detection-segmentation networks.

Detection networks are designed to locate, classify and label existing objects in any image using a bounding box. 
This is a~set of coordinates of the corners of the rectangles that mark the detected objects in the image. 
A conventional method of object detection is based on proposing regions and then classifying each proposal into different object categories. 
This includes network architectures based on regions with convolutional neural networks (R-CNN) \cite{rcnn}.
Another approach considers object detection as a regression or classification problem in order to directly obtain the final results (categories and locations).
These include, among others, the YOLOv7 network architectures \cite{yolov7}.

Segmentation networks are based on an encoder-decoder architecture. 
They are used to classify each pixel in the image.
Two types of segmentation can be distinguished: semantic and instance.
A~representative example of semantic segmentation is U-Net \cite{unet}. 
The encoder module uses convolution and pooling layers to perform feature extraction. 
On the other hand, the decoder module recovers spatial details from the sub-resolution features, while predicting the object labels.
A~standard choice for the encoder module is a lightweight CNN backbone, such as GoogLeNet 
or a revised version of it, namely Inception-v3 \cite{inceptionv3}. 
%
%
To improve the accuracy and efficiency of semantic segmentation networks, multi-branch architectures have been proposed. 
They allow high-resolution segmentation of objects in the image.
To this end, multi-branch networks introduce a fusion module to combine the output of the encoding branches. 
This can be a Feature Fusion module in which the output features are joined by concatenation or addition, an Aggregation Layer (BiSeNet V2 \cite{BiSeNetv2}), a Bilateral Fusion module (DDRNet \cite{DDRNet}) or a Cascade Feature Fusion Unit (ICNet \cite{ICNet}).
\add{Moreover, there is more and more research in the direction of object detection and segmentation to use transformer-based neural networks, such as DETR}\cite{DETR}\add{,  SegFormer}\cite{SegFormer}. \add{In the segmentation task, there are only a few architectures proposed at the moment, while in the object detection task there are many solutions, of which transformer-based methods achieve the best performance. Vision transformers offer robust, unified, and even simpler solutions for various tasks. Compared to CNN approaches, most transformer-based approaches have simpler pipelines but stronger performance. However, transformer-based methods require a lot of training data.}

Many dedicated solutions require both detection and segmentation of objects in the image.
It should be noted that once full segmentation (i.e. for all object classes under consideration) has been performed, there is no need to implement detection -- the bounding boxes can be obtained from the masks of individual objects.
However, networks implementing accurate multi-class semantic segmentation or instance segmentation are characterized by high computational complexity, as highlighted in the paper \cite{benchmarks}. 
The authors showed that the performance of the three most accurate networks did not exceed 24 fps (frames per second) on an RTX 3090 and 12 fps on a GTX 1080 Ti graphics card.
This shows that for this type of network, achieving real-time processing (60 fps) on an embedded computing platform is challenging. 
Hence the idea of combining fast detection with segmentation limited to a few classes with relatively little variation (such as roadway, road markings, or vegetation/buildings).
A~key feature of this type of solution is the encoder common to both functionalities.
This approach makes it possible to run deep networks on embedded devices equipped with computing chips that consume less power but have also less computing power.
Furthermore, as will be shown later, the process of learning a segmentation-detection network is easier and faster than, an alternative solution based on a segmentation network only.
In the papers \cite{multinet, joint_detection_segmentation, dlt_net, multitask, HybridNet, YOLOP}, detection-segmentation network architectures have been proposed that currently achieve the best results. 
The training process typically uses the following datasets: \textit{KITTI}, \textit{Cityscapes}, VOC2012 or \textit{BDD100k} \cite{kitty, inproceedings_cityscapes, voc2012, bdd100k}.

When pre-selecting the appropriate solutions for the experiments, we took into account the diversity of the proposed architectures, the fulfillment of the requirements related to the FPT'22 competition \cite{FPT}, as well as the possibility of quantizing and accelerating the network on embedded computing platforms, i.e. eGPU (embedded Graphic Processing Unit), SoC FPGA (System on Chip Field Programmable Gate Array).
Therefore, we decided to use the following three networks in our research: MultiTask V3 \cite{multitask}, HybridNet \cite{HybridNet}, and YOLOP \cite{YOLOP}.

\section{The considered detection-segmentation neural networks}
\label{sec:det_seg_nn}

\begin{figure}[!]
\centerline{\includegraphics[width=0.49\textwidth]{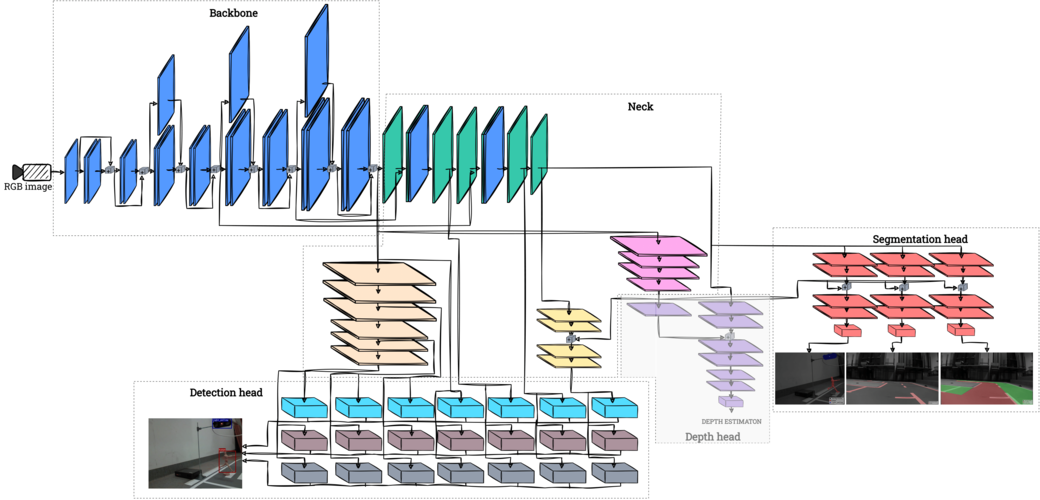}}
\caption{Scheme of the MultiTask V3 neural network architecture}
\label{fig:multitask_scheme}
\end{figure}

The MultiTask V3 network \cite{multitask} is a model proposed by the developers of the Vitis AI (AMD Xilinx) platform for users using neural networks on SoC FPGA platforms. 
A scheme of the MultiTask V3 neural network architecture is shown in Figure \ref{fig:multitask_scheme}.
It allows five tasks to be performed simultaneously -- detection, three independent image segmentations, and depth estimation. The backbone of the network, which determines the underlying feature vector, is based on the ResNet-18 convolutional neural network. 
Subsequent features are extracted using encoders and convolutional layer blocks. 
Branches responsible for a given part of the network then generate the corresponding output using convolution, ReLU activation operations, and normalization.
Due to the large number of tasks to be performed, the network was trained to segment road markings, lanes (including direction), and objects (pedestrians, obstacles, and traffic lights) separately. 
Detection was performed on the same set of objects.
The model was trained using only our own custom datasets, which were transformed into the format recommended by the network developers. 
The resulting network processes images with a~resolution of $512 \times 320$ pixels. 
In addition, thanks to the model quantization tools, it is possible to reduce the precision and run the network on SoC FPGA platforms using DPUs (Deep Learning Processor Units). 
The performance on the original \textit{BDD100k} dataset \cite{bdd100k} was not given, as the network was not previously described in any scientific paper.

\begin{figure}[!]
\centerline{\includegraphics[width=0.49\textwidth]{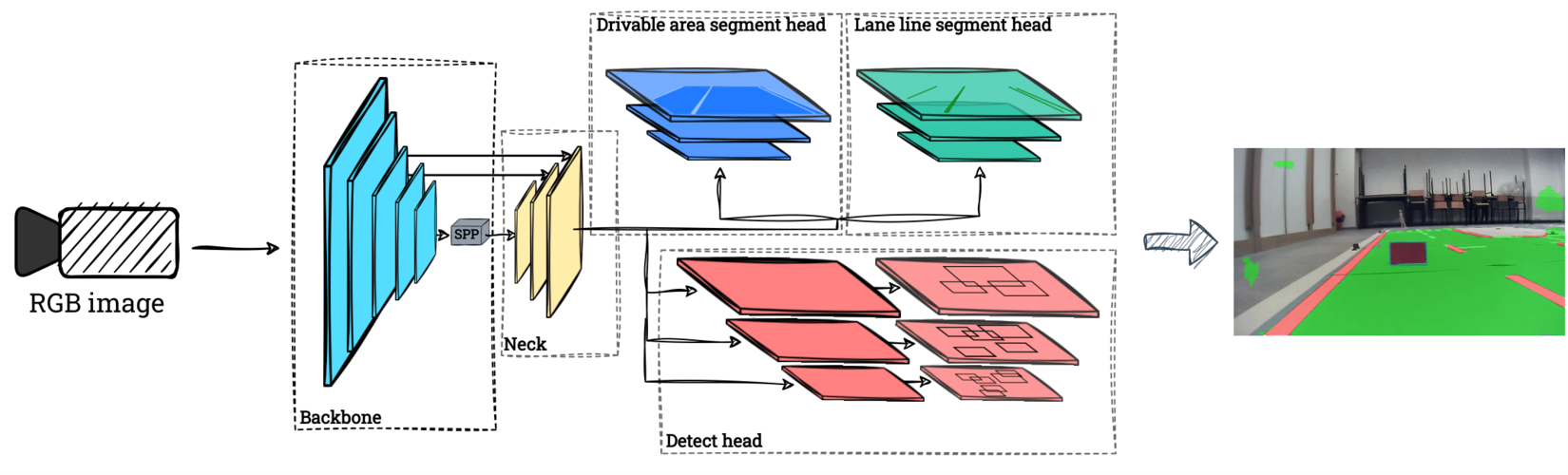}}
\caption{Scheme of the YOLOP neural network architecture.}
\label{fig:yolop_scheme}
\end{figure}

The second detection-segmentation neural network considered is the YOLOP \cite{YOLOP}.
A scheme of the architecture is shown in Figure \ref{fig:yolop_scheme}.
It performs 3 separate tasks within a~single architecture -- detection of objects in the road scene, segmentation of the drivable area, and road markings. 
The network consists of a~common encoder and 3 decoders, with each decoder dedicated to a separate task. 
The drivable area represents all lanes in which the vehicle was allowed to move -- opposite lanes were not taken into account.
The network was originally trained on the \textit{BDD100k} dataset \cite{bdd100k}.
To reduce memory requirements, the images were scaled from a resolution of $1280 \times 720 \times 3$ to a resolution of $640 \times 384 \times 3$.
The network achieved a $mAP_{50}$ score for single class detection (cars) of 76.5\%, drivable area segmentation mIoU of 91.5\%, and lane line segmentation mIoU score of 70.5\%. 

\begin{figure}[!]
\centerline{\includegraphics[width=0.4\textwidth]{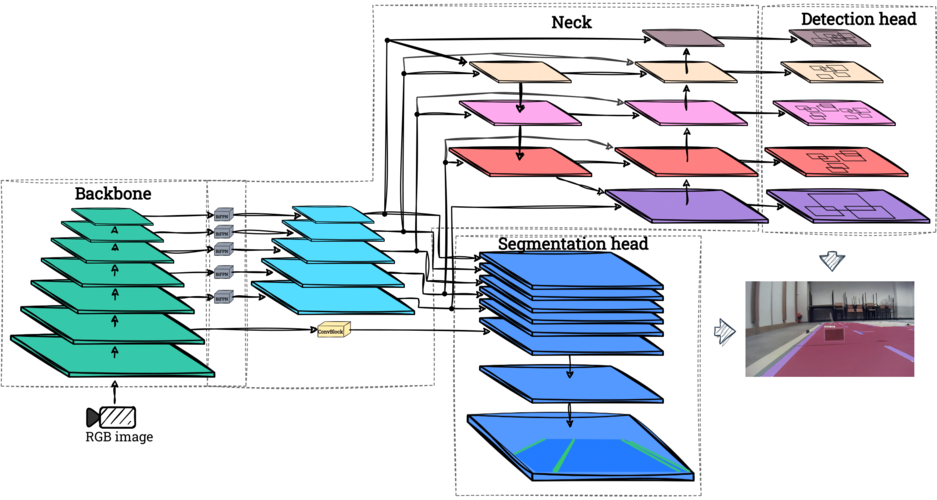}}
\caption{Scheme of the HybridNets neural network architecture.}
\label{fig:hybridnet_scheme}
\end{figure}

HybridNets \cite{HybridNet} network is another example of simultaneous segmentation and detection models. 
HybridNets, like YOLOP, only performs object detection and segmentation of road markings and drivable area (without considering lane direction). 
A scheme of the architecture is shown in Figure \ref{fig:hybridnet_scheme}.
It does not have the semantic segmentation and depth estimation branches available in MultiTask V3.
The network consists of four elements: a feature extractor in the form of EfficientNet V2 \cite{EfficientNetV2}, a neck in the form of BiFPN \cite{BiFPN}, and two branches, one for a detection head similar to YOLOv4 \cite{YOLOv4} and the other for segmentation consisting of a series of convolutions and fusion of the outputs of successive layers of the neck. 
The network was initially trained on the \textit{BDD100k} dataset \cite{bdd100k}, whose images were scaled to a size of $640 \times 384 \times 3$.
It achieved a $mAP_{50}$ for single class detection (cars) equal to 77.3\%, drivable area segmentation mIoU of 90.5\%, and lane line segmentation mIoU score of 31.6\%.

\section{Experiments performed}
\label{sec:experiments}



\begin{figure}
\begin{tabular}{cccc}

\subfloat[Object detection]{\includegraphics[width = 1.5in]{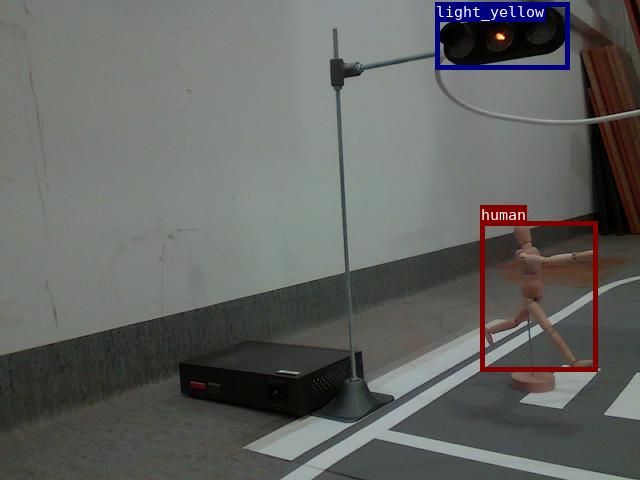}} &
\subfloat[Object segmentation]{\includegraphics[width = 1.5in]{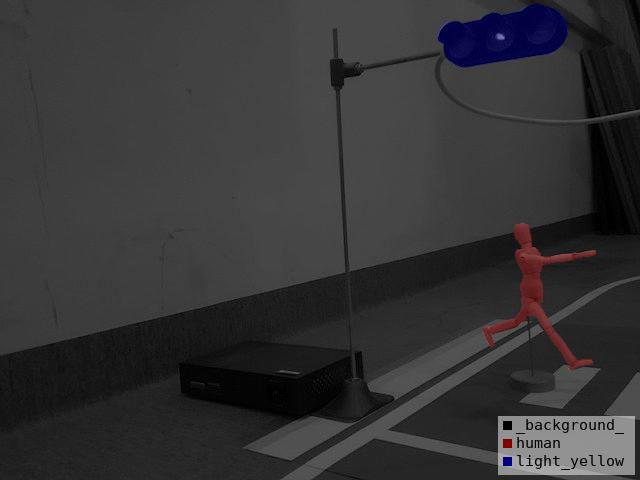}} \\
\subfloat[Drivable area segmentation]{\includegraphics[width = 1.5in]{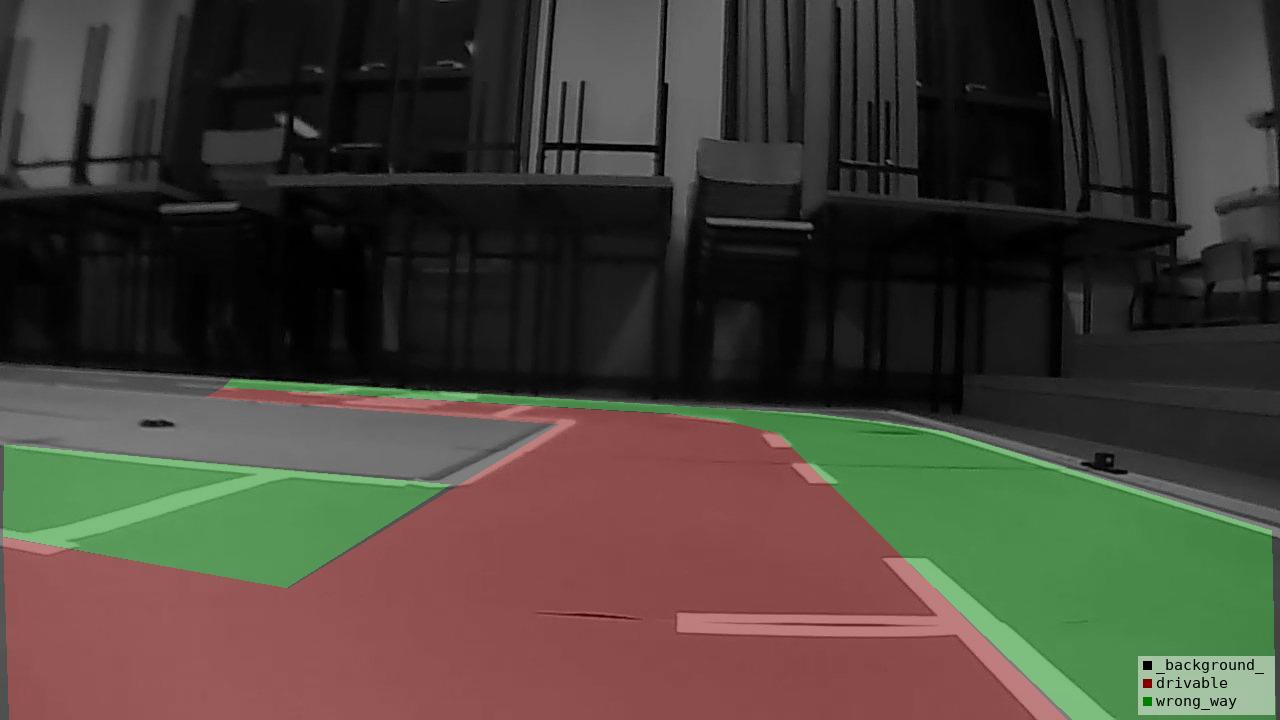}} &
\subfloat[Lane segmentation]{\includegraphics[width = 1.5in]{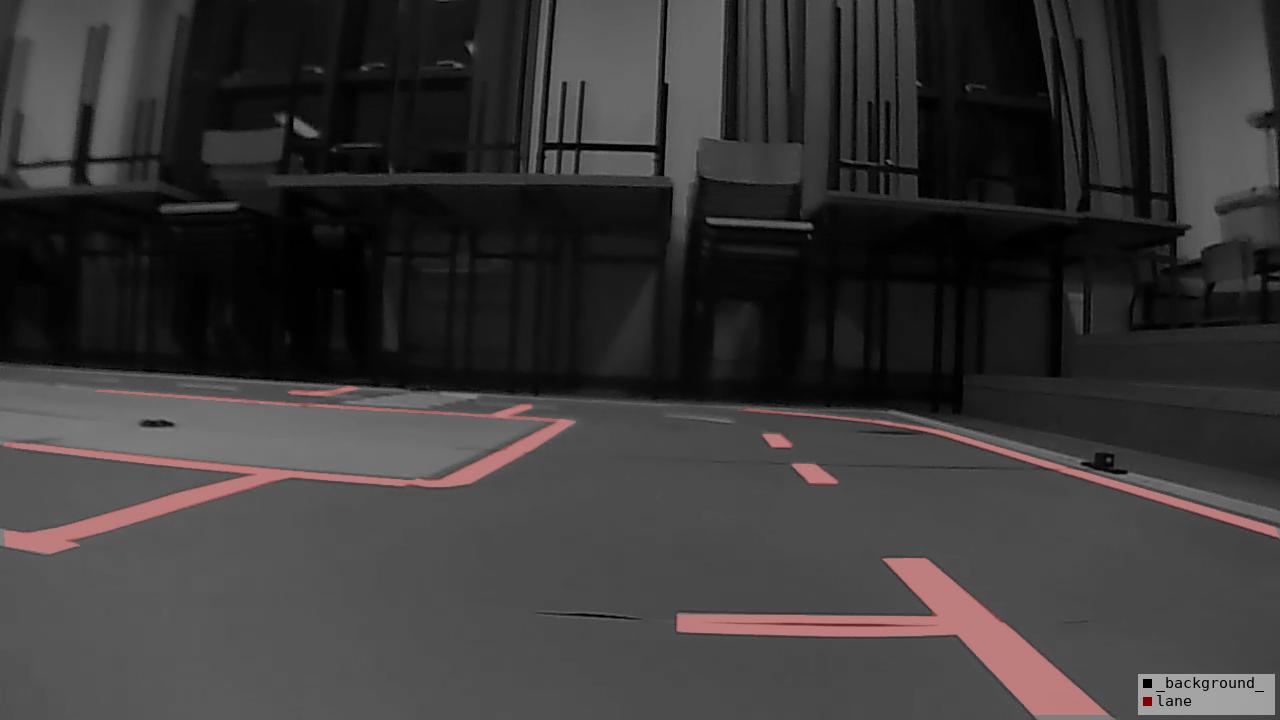}} 

\end{tabular}
\caption{Examples of training sets. Set (b) was generated for the MultiTask V3 network only, and sets (a), (c) and (d) for all models.}
\label{fig:example_dataset}
\end{figure}

A custom training dataset was prepared to compare the above-mentioned neural network models.
It was divided into three subsets containing objects (pedestrian figures, obstacles, traffic lights), road markings, and drivable area, respectively.
The subsets were prepared based on the collected recordings from the city mock-up which was constructed according to the rules of the FPT'22 competition \cite{FPT}.
Subsequently, labels were applied to the images obtained from the recordings.
The road markings dataset was prepared semi-automatically by pre-selecting a threshold and performing binarization. 
Annotations were then prepared for all sets using the Labelme software \cite{labelme}.
The resulting label sets were adapted to the formats required by the tools designed to train the aforementioned networks. 
The final dataset consisted of 500 images of the city mock-up with road markings, 500 images with the drivable area, and 250 images with objects.
\add{Size of the dataset was dictated by a small environment with little changes of lightning and camera angles, as the purpose of the trained model was to be used only on a given mock-up.}
The prepared datasets were divided into training and validation subsets in an 80/20 ratio.
\add{The validation set was later used as the test set. This decision was made because the size of the prepared dataset was relatively small (but still sufficient to properly train the model, as shown in Figure} \ref{fig:comparison1}.
An example of an input data set from a training set is shown in Figure \ref{fig:example_dataset}.


In the case of the MultiTask V3 network, a path to the prepared dataset was passed to the training program. 
The application managed the training sets independently so that it was possible to run the training procedure from start to finish on all sets.
\add{Network has been trained using default hyperparameters, provided by developers. The base learning rate was set to 0.01. The optimiser used for training was a Stochastic Gradient Descent (SGD). Training included data augmentation as random mirroring of the input images, photometric distortion, and random image cropping. The model was trained using a batch size of 16.}
As the MultiTask V3 network also performs object segmentation, the maximum number of epochs was set to the highest of all the models considered. A value of 450 epochs was chosen, after which no significant increase in validation results was observed.

The YOLOP network training program did not allow different parts of the model to be trained simultaneously with independent sets.
As the segmentation sets were different from the detection set, it was necessary to split the network training procedure.
The training procedure began with the backbone layers and the upper detection layers (segmentation layers were frozen). Once this was completed, the layers responsible for segmentation were unfrozen, the remaining layers were frozen, and the training procedure was restarted.
\add{Network has been trained using default hyperparameters, provided by developers. 
The base learning rate was set to 0.01. The optimiser used for training was an Adam algorithm. 
Training included data augmentation as random changes in image perspective and random changes in the image's colour hue, saturation, and value. The model was trained using a batch size of 2.}
The training was stopped after 390 epochs, as the validation results did not improve in the following steps.

As with YOLOP, the HybridNet training program does not allow simultaneous training with two independent data sets.
Therefore, a similar training strategy to YOLOP was used.
First, the backbone and the detection branch were trained.
\add{Default hyperparameter settings provided by the developers were used, including the AdamW optimiser. 
There were only two parameters that were changed: a batch size of 4 and an initial learning rate of 0.001.
The change of learning rate during detection training was chosen, as starting with the default learning rate of 0.0001 didn't show promising results.}
After 150 epochs, when no further performance improvement was observed on the validation set, training was stopped, the backbone and detection branches were frozen, and training was started on the segmentation set.
\add{This time the default hyperparameters were kept, including the learning rate of 0.0001.}
The segmentation branch was trained for 100 epochs until no improvement in performance was observed.
In total, the network was trained for 250 epochs.
\add{Both of the branches were trained using the default data augmentation provided by the researchers, in the form of: left-right flip, change of hue, rotation, shear and translation.}

\section{Results and Discussion}
\label{sec:results}



\begin{figure}[!]
\begin{tabular}{cc}

\subfloat[Original image]{\includegraphics[width = 1.5in]{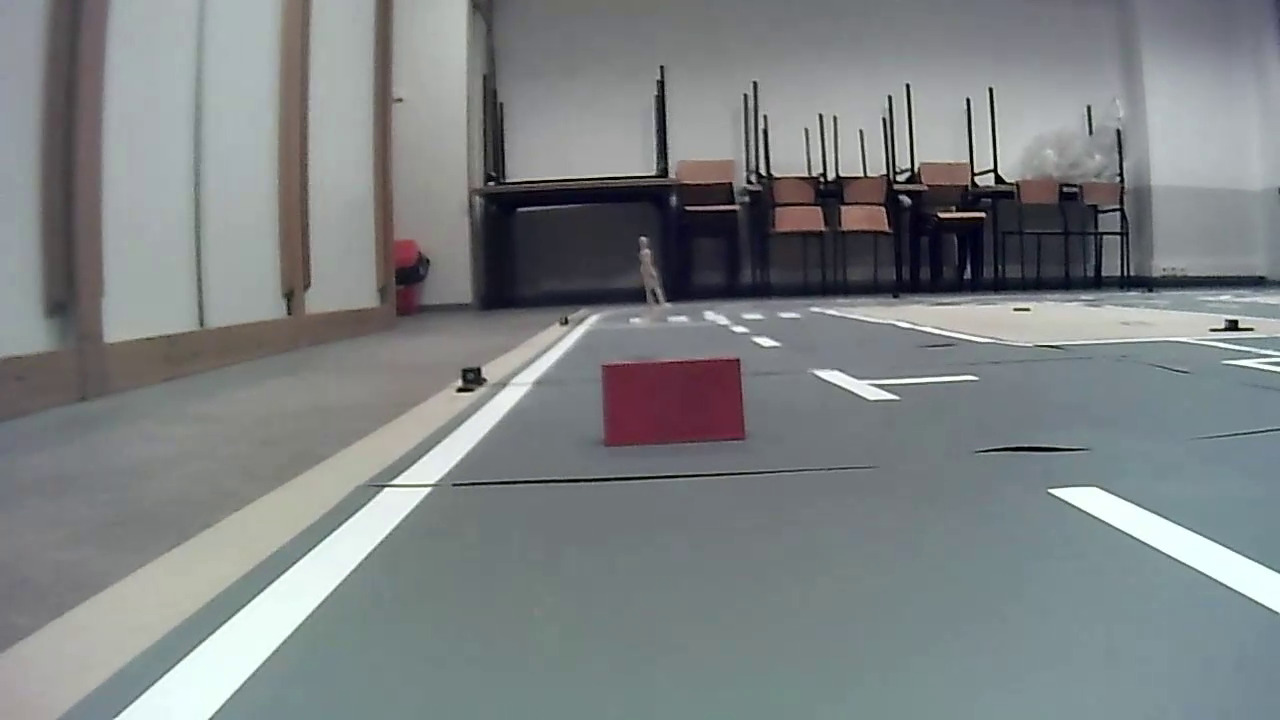}} &
\subfloat[Result of the YOLOP]{\includegraphics[width = 1.5in]{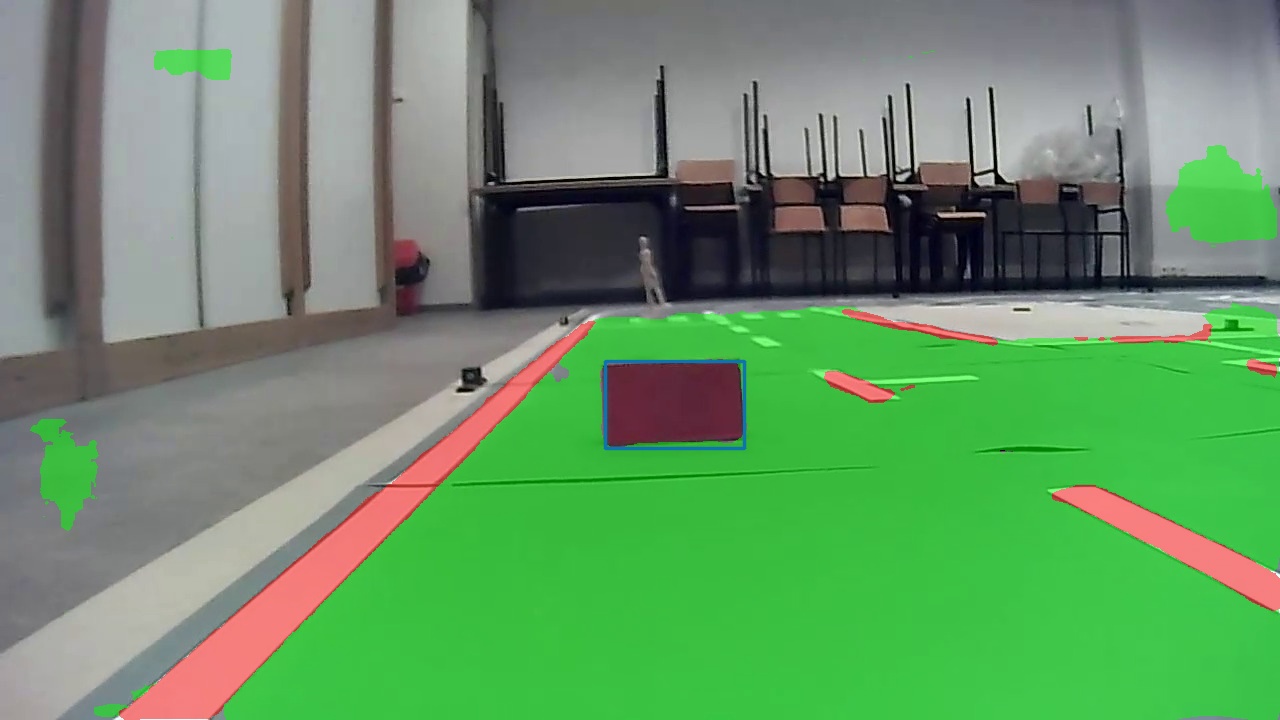}}\\
\subfloat[Result of the MultiTask V3]{\includegraphics[width = 1.5in]{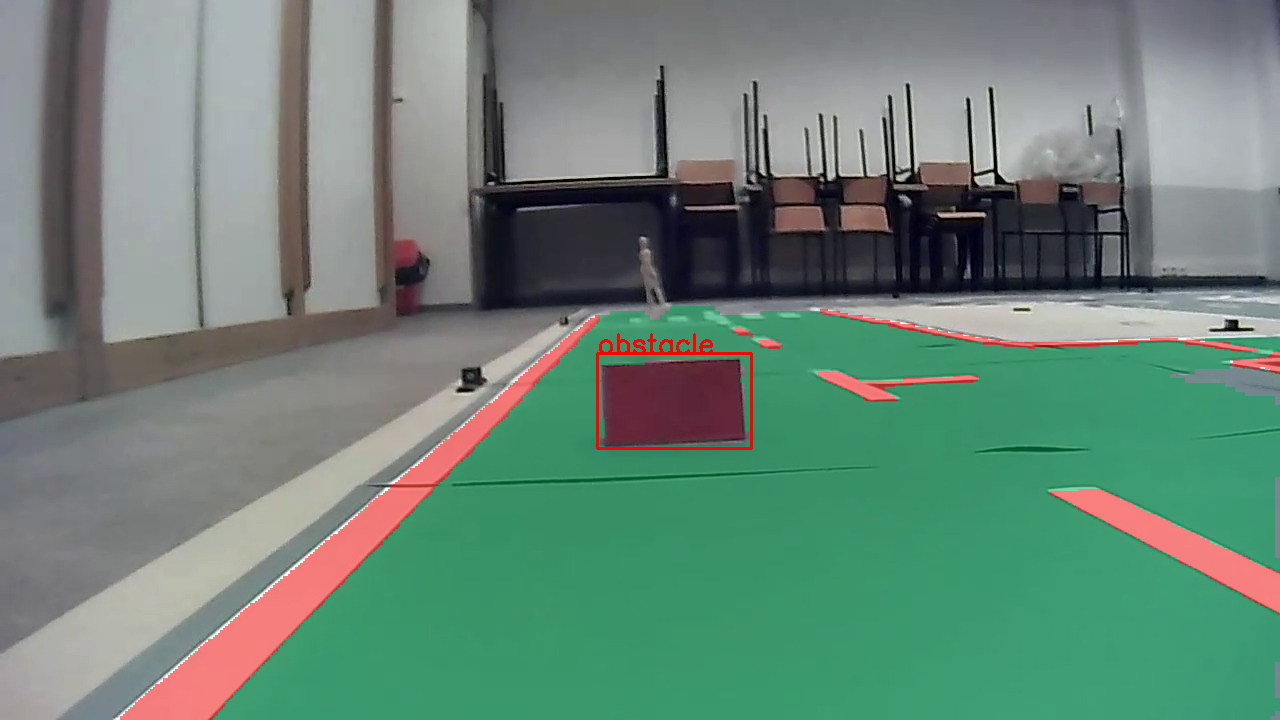}} &
\subfloat[Result of the HybridNet]{\includegraphics[width = 1.5in]{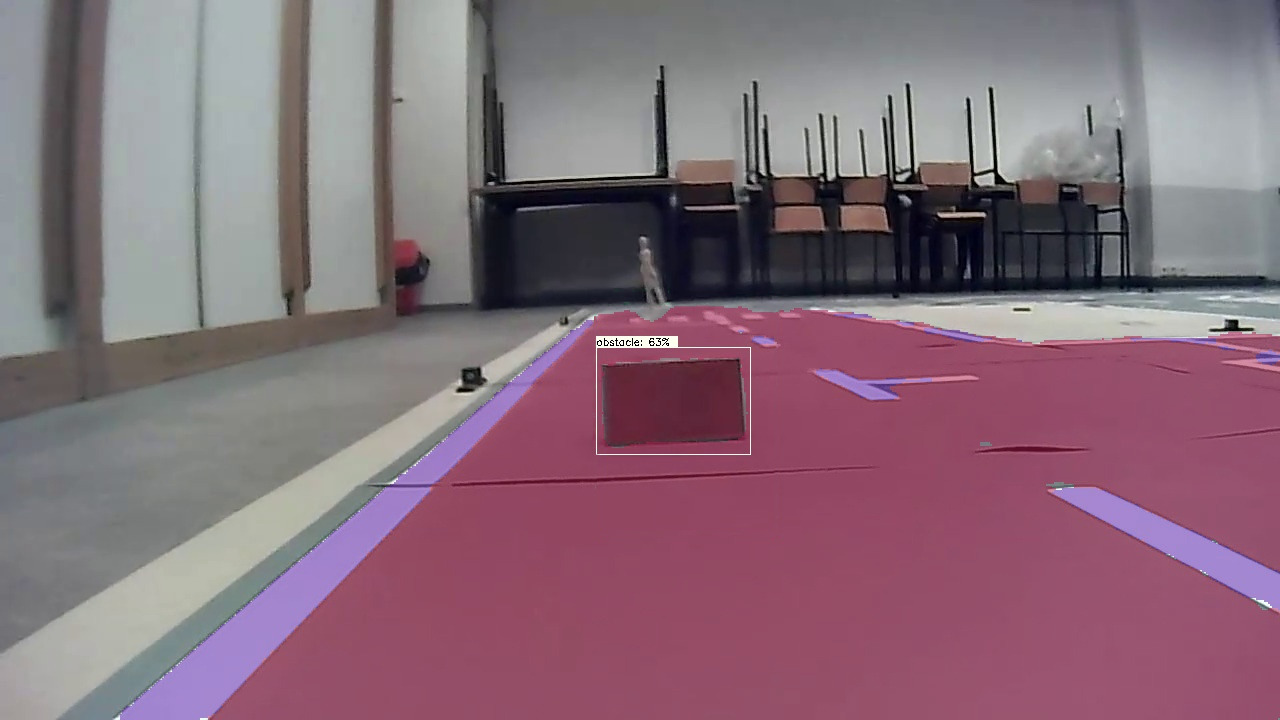}} 

\end{tabular}
\caption{Comparison of network performance on a sample including an object on a straight road.}
\label{fig:comparison1}
\end{figure}

Figure \ref{fig:comparison1} shows the results of the considered neural network in terms of object detection, driving area, and road marking segmentation for a view containing a straight road.
To verify the effectiveness of our selected detection-segmentation neural network models, we compared the performance of each single-task scheme separately, as well as the multitask scheme.

\begin{table}[h!t]
\centering
\caption{Comparison of performance on the GTX 1060 M and RTX 3060 graphics cards.}
\begin{tabular}{|c|c|c|c|c|c|}
\hline

\multirow{2}{*}{GPU} & \multirow{2}{*}{Model} &   Speed &  Inference \\
{} &{} & [fps] &  time [s] \\
\hline
\multirow{3}{*}{GTX 1060 M} & YOLOP   & 45.05 & 0.0222\\
{} & MultiTask V3  &  75.55 & 0.0132\\
{} & HybridNets  & 7.35 & 0.1361\\
\hline
\multirow{3}{*}{RTX 3060} & YOLOP    & 123.45 & 0.0081\\
{} & MultiTask V3   &   124.34 & 0.0080\\
{} & HybridNets   &  27.39 & 0.0365\\
\hline
\end{tabular}
\label{tab:efficiency}
\end{table}

Table \ref{tab:efficiency} shows the performance of the models on the NVIDIA GeForce GTX 1060 M and NVIDIA GeForce RTX 3060 graphics cards.
It can be seen that YOLOP and MultiTasks networks for comparable resolutions process data in real-time, while HybridNets is slightly slower.
Here it should be noted that the original implementation of HybridNets was used.
Unlike the YOLOP and MultiTask V3 models, it makes extensive use of subclassing to implement most of the layers used in the network.
This may cause large discrepancies in the inference speed of the network compared to other models.
Table \ref{tab:complexity} summarises the input image resolution and computational complexity of the selected neural networks.
MultiTask V3 has the highest FLOPS value, especially when normalised with respect to the input image resolution and the highest number of parameters.
On the other hand, it achieved the best performance on both GPUs, possibly due to the highly optimised parallel implementation. 
We then performed an evaluation to assess the performance of each task: object detection and drivable and lane segmentation.
We considered the object detection performance of three models on a custom dataset. 
As shown in Table \ref{tab:detection}, we use $mAP_{50}$, $mAP_{70}$, and $mAP_{75}$ as the evaluation metrics of the detection accuracy.
For YOLOP and MultiTask V3, the $mAP_{50}$ score is above 95\%, proving that both networks have been successfully trained. 
For MultiTask V3, the score does not change much as the IoU (Intersection over Union) acceptance level increases, while for YOLOP it decreases slightly. 
This result shows that the detections made by MultiTask V3 are very similar to those provided by the validation dataset, while YOLOP's detections are close to them, but do not overlap perfectly. 
The $mAP_{50}$ score for the HybridNets architecture is about 84\%.
This score is lower than the previous two architectures but still allows for acceptable detection accuracy.
We used IoU and mIoU (mean IoU) as evaluation metrics for drivable area segmentation and lane segmentation accuracy.
A comparison of the drivable area segmentation results for MultiTask V3, YOLOP and HybridNets is shown in Table \ref{tab:drivable_results}. 
Note that one of the requirements of the FPT'22 competition is left-hand traffic.
It can be seen that the best performance is achieved by the MultiTask V3 network.
However, the other neural networks also perform very well, with an accuracy of no less than 84\%. 
A high IoU score for the drivable area class for all networks shows that the predicted segmentations are almost the same as those in the validation dataset. 
Achieving such high results was predictable as the driving area surfaces are large, simple in shape, and uniform in color. 
It is therefore relatively easy to distinguish them from the background.
As the background is classified as any other pixel not belonging to the driving area class, the results obtained are even higher.

The results of the\change{evaluation of}{lane segmentation evaluation} \add{are shown in Table} \ref{tab:lane_results}.
There are much lower than for the drivable area segmentation, which can be expected, as lane markings are much smaller than the drivable area's planes. 
Their shape is more complex as well.
The values of Lane IoU for each of the neural networks vary between about 79 to 91\%, the best for MultiTask V3.
Here it should be noted, that the quality metrics for the HybridNets model are about 10 percentage points lower for each task.
Many approaches were tried by changing the hyperparameters, such as batch size, optimiser, learning rate, backbone size, training order, and even transfer learning approaches, but none gave better results.
This could be due to the lack of the final step of the proposed learning algorithm, where the optimiser is changed to SGD \remove{(Stochastic Gradient Descent)} and both tasks and the backbone are trained at the same time.
HybridNets may be more receptive to simultaneous learning due to the greater influence of the shared neck, as its architecture is highly interconnected.
Further ablation studies would be needed to determine the reasons for the observed lower performance.

To sum up conducted experiments, it should be noted that the best results in each category were obtained for the MultiTask V3 network. 
However, it has a certain disadvantage in terms of computational and especially memory complexity (the highest of the networks considered). 
On the other hand, it works in real-time on both GPU cards considered. 
In addition, it also allows for the acquisition of depth information (not considered in this study). 
Taking all these factors into account, the MultiTask V3 network should be considered a very good candidate for building an embedded perception system for an autonomous vehicle. 
The code used in the described experiments is available at: \textit{removed for blind revision}.
The code used in the described experiments is available at: \url{https://github.com/vision-agh/MMAR_2023}.






\begin{table}[!t]
\centering
\caption{Comparison of computational complexity.}
\begin{tabular}{|c|c|c|c|c|} 
\hline
\multirow{2}{*}{Model} & Resolution & FLOPS & Normalized & Params \\
{} & [px] & [G] & FLOPS [K/px]& [M]\\
\hline
YOLOP & $640 \times 640$ & 17.32 & 44.32 & 7.94 \\
MultiTask V3 & $512 \times 320$ & 25.44 & 162.82 & 16.43 \\
HybridNets & $640  \times 384$ & 14.53 & 57.30 & 13.43 \\
\hline
\end{tabular}
\label{tab:complexity}
\end{table}

\begin{table}[!t]
\centering
\caption{Comparison of results for object detection.}
\begin{tabular}{|c|c|c|c|}
\hline
Model &  $mAP_{50}$ [\%] &  $mAP_{70}$ [\%] &  $mAP_{75}$ [\%]\\
\hline
YOLOP   & 96.2 & 86.6 & 75.5\\
MultiTask V3   &  99.4 & 99.4 & 97.2\\
HybridNets   & 83.3 & 79.7 & 78.5\\
\hline
\end{tabular}
\label{tab:detection}
\end{table}

\begin{table}[!t]
\centering
\caption{Comparison of results for drivable area segmentation.}
\begin{tabular}{|c|c|c|c|} 
\hline
\multirow{2}{*}{Model} & \multirow{2}{*}{MIoU [\%]} & \multicolumn{2}{c|}{IoU [\%]}  \\
 \cline{3-4}
 &  & Background & Drivable area \\ 
\hline
YOLOP & 92.50 & 94.00 & 91.00\\
MultiTask V3 & 97.28 & 97.86 & 96.70 \\
HybridNets & 88.3 & 91,7 & 84.9\\
\hline
\end{tabular}
\label{tab:drivable_results}
\end{table}

\begin{table}[!t]
\centering
\caption{Comparison of results for lane segmentation.}
\begin{tabular}{|c|c|c|c|} 
\hline
\multirow{2}{*}{Model} & \multirow{2}{*}{MIoU [\%]} & \multicolumn{2}{c|}{IoU [\%]}  \\
\cline{3-4}
 &  & Background & Lanes \\ 
\hline
YOLOP & 86.20 & 98.50 & 73.90 \\
MultiTask V3 & 91.06 & 99.08 & 83.03 \\
HybridNets & 79.1 & 91.9 & 66.3 \\
\hline
\end{tabular}
\label{tab:lane_results}
\end{table}

\section{Conclusion}
\label{sec:conclusion}

In this paper, we compared three detection-segmentation convolutional neural network architectures: MultiTask V3, YOLOP, and HybridNets.
We used a custom dataset prepared according to the requirements of the FPT'22 competition.
\add{The dataset was created solely for training models that could be used on the city mock-up. Due to the constant environmental factors and relatively few corner cases (such as intersections, turns, etc.), there was no need to obtain more data. However, for real-world applications, more work should be done to prepare the dataset. It should include more data, including different locations and environments (lighting, weather factors, etc.) to make the models reliable in a diverse environment.}
The results obtained confirm the high attractiveness of this type of networks -- they allow good detection and segmentation accuracy, and real-time performance.
Moreover, the training of these networks is simpler, since certain parts can be trained independently, even on separate datasets.
Of the three methods analysed, MultiTask V3 proved to be the best, obtaining 99\% $mAP_{50}$ for detection and 97\% MIoU for drivable area segmentation and 91\% MIoU for lane segmentation, as well as 124 fps on the RTX 3060 graphic card.
This architecture is a good solution for embedded perception systems for autonomous vehicles.
As part of future work, we plan to focus on two further stages of building an embedded perception system based on a deep convolutional neural network.
First, we want to perform quantization and pruning of the analysed network architectures to see how they will affect efficiency and computational complexity.
Next, we will run the networks on an eGPU (e.g. Jetson Nano) and an SoC FPGA (\remove{System on Chip Field Programmable Gate Array,}e.g. Kria from AMD Xilinx).
\add{Networks will be compared on given platforms for performance and power consumption. It is worth noting that initial tests on an eGPU with MultiTask V3 and YOLOP have shown, showing that MultiTask V3 provides faster inference while consuming less energy.}
In the final step, we will add a control system to the selected perception system, place the selected computational system on a model of an autonomous vehicle and test its performance on the created mock-up.
Secondly, we will attempt to use the \textit{weakly supervised learning} and \textit{self-supervised learning} methods, which, in the case of an atypical, custom dataset, would allow a~significant reduction in the labeling process of the learning data.
Thirdly, we also want to consider adding modules for depth estimation and optical flow, as elements often used in autonomous vehicle perception systems. 






\section*{Acknowledgment}
The work presented in this paper was supported by the AGH University of Krakow project no. 16.16.120.773 and by the programme “Excellence initiative – research university” for the AGH University of Krakow.

\vspace{12pt}

\end{document}